\documentclass{article}

\usepackage[preprint]{neurips_2022}

\usepackage[utf8]{inputenc} 
\usepackage[T1]{fontenc}    
\usepackage{hyperref}       
\usepackage{url}            
\usepackage{booktabs}       
\usepackage{amsfonts}       
\usepackage{nicefrac}       
\usepackage{microtype}      
\usepackage{xcolor}         
\usepackage[pdftex]{graphicx}
\usepackage{natbib}
\setcitestyle{numbers,square}
\def\eg{\emph{e.g.}}

\def\etal{\emph{et al.}}
\def\etc{\emph{etc.}}
\def\ie{\emph{i.e.}}

\title{Dress Well via Fashion Cognitive Learning}

\usepackage{authblk}

\author{Kaicheng Pang}
\author{Xingxing Zou}
\author{Waikeung Wong\thanks{The corresponding author.}}
\affil{School of Fashion and Textiles, The Hong Kong Polytechnic University}
\affil{Laboratory for Artificial Intelligence in Design}
\affil{\texttt{\{kaicpang.pang, aemika.zou\}@connect.polyu.hk, calvin.wong@polyu.edu.hk}}

\begin{document}

\maketitle
\begin{abstract}
   Fashion compatibility models enable online retailers to easily obtain a large number of outfit compositions with good quality. However, effective fashion recommendation demands precise service for each customer with a deeper cognition of fashion. In this paper, we conduct the first study on fashion cognitive learning, which is fashion recommendations conditioned on personal physical information. To this end, we propose a Fashion Cognitive Network (FCN) to learn the relationships among visual-semantic embedding of outfit composition and appearance features of individuals. FCN contains two submodules, namely outfit encoder and Multi-label Graph Neural Network (ML-GCN). The outfit encoder uses a convolutional layer to encode an outfit into an outfit embedding. The latter module learns label classifiers via stacked GCN. We conducted extensive experiments on the newly collected O4U dataset, and the results provide strong qualitative and quantitative evidence that our framework outperforms alternative methods.
\end{abstract}

\section{Introduction}
\begin{figure}[t]
   \begin{center}
   \includegraphics[width=1\linewidth]{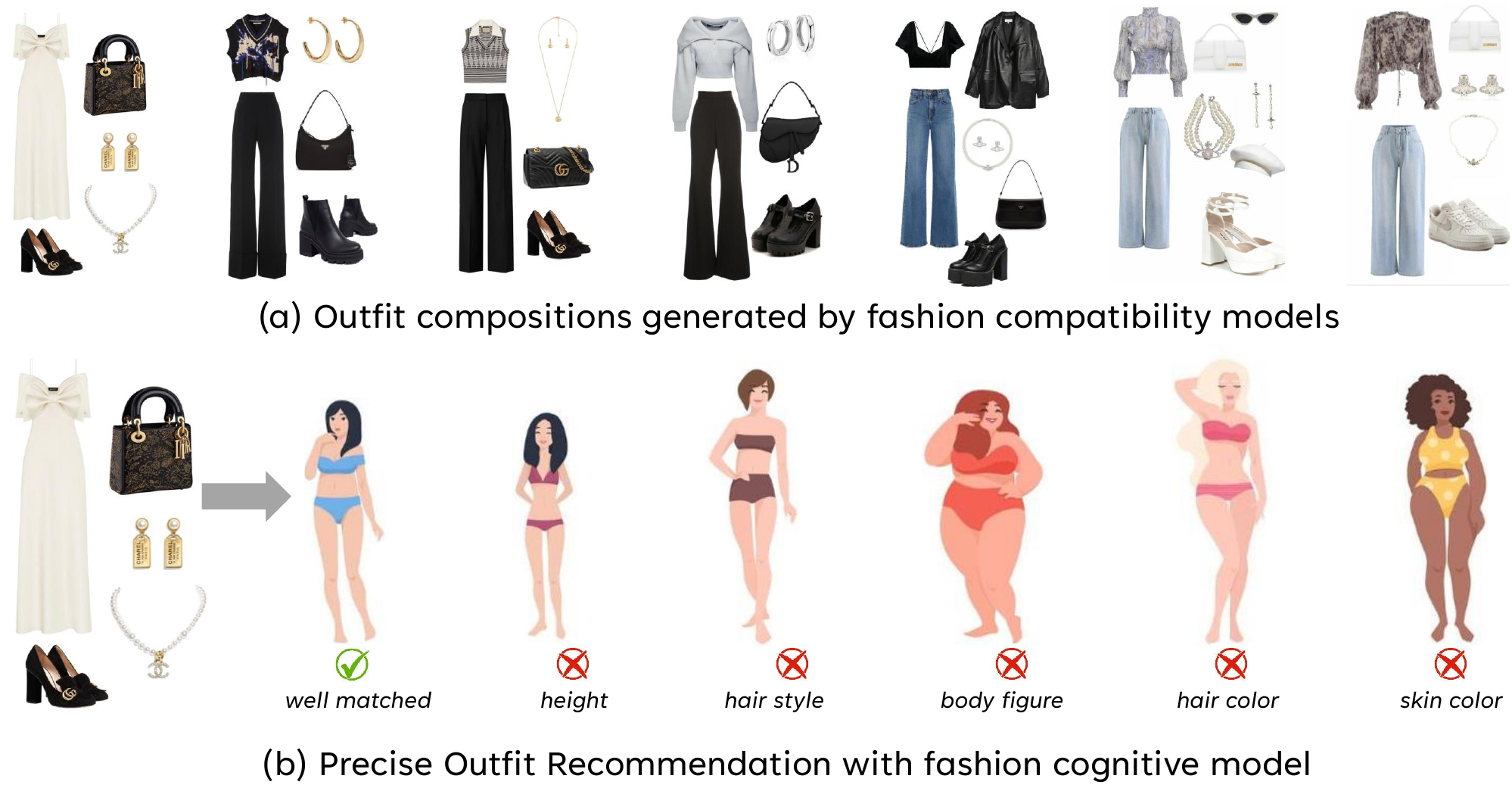}
   \vspace{-1cm}
   \end{center}
   \caption{Current situations occur in fashion online retailing.}
   \label{1}
   \vspace{-0.5cm}
\end{figure}

Fashion exists~\citep{nakamura2018outfit} in our daily life as a tool for expressing attitude and presenting culture. However, it is a problem when applying it to online products since the fashion compatibility models only solved the task that fashion items in an outfit are compatible with each other but have not considered whether the outfits are compatible with customers when they are shopping online.
As shown in Figure~\ref{1} (b), different customers have varied appearances, \eg, different heights, hairstyles, skin colors, \etc, which will directly affect whether an outfit is compatible with them or not. For example, the outfit consists of a white long dress that is not suitable for the second customer since she is not so high enough to wear this long dress. Thus, even though this outfit itself is perfectly matched, it is not appropriate to recommend it to her. Otherwise, it will be resulting she losing trust in the service provider. In other words, understanding relationships between outfits and customers to achieve precise outfit recommendations is crucial.

Previous research mainly focused on the relations among fashion items via fashion compatibility learning~\citep{han2017learning,wang2019outfit,zhang2020learning,lin2020fashion}. Many of them~\citep{zou2020regularizing,lin2018explainable} also focused on the explainability of fashion compatibility models. In addition, a few works noticed the influence of personal information, \eg user preference~\citep{chen2019pog,liu2018personal,packer2018visually}, social media posts~\citep{zheng2021personalized,sonie2019personalised}, body shape~\citep{hidayati2018dress}. However, no prior approach systematically considered the compatibility relationships among fashion items in an outfit and the varied appearance of online shoppers and then solved it from a comprehensive point of view.

In this work, we specifically target the task aiming to provide precise and appropriate fashion recommendation service for each customer. To distinguish from previous works utilizing the user's personal preference for personalized recommendation, we define the new task as \textbf{Fashion Cognitive Learning}, \ie, focusing on the influence of personal physical information on the compatibility of an outfit. To achieve this, we treat this task as a multi-label classification task and propose an end-to-end framework, namely Fashion Cognitive Network, that learns the compatible relationships between outfits and humans. The FHN first uses several convolutional filters with different window sizes to encode the outfit into an outfit embedding. Using filters of different sizes enables convolutional kernels to see different combinations of fashion attribute features. Then the text of each physical label will be embedded into a word embedding and then go through a stacked GCN to learn multi-label classifiers. Finally, the predicted scores for all physical labels are obtained by multiplying classifier vectors with outfit embedding. 

Meanwhile, to facilitate the development of our framework, we introduce a new outfit dataset covering personal physical information, namely Outfits for You (O4U). It includes a total of 29352 outfits. Each outfit has two types of label: 1. this outfit is good or not; 2. this outfit is not compatible with which kind of physical label. We invited several fashion experts to label these outfits. The experiment were conducted on the O4U dataset and compare the proposed FCN with state-of-the-art methods. The main contributions of this work are summarized as follows:

\begin{itemize}
   \item We introduce a new task, i.e., fashion cognitive learning, which targets learning the compatibility relationships between outfits and personal physical information in an end-to-end framework to facilitate precise fashion recommendations.
   \item We introduce a new outfit dataset with tremendous personal physical information for facilitating fashion cognitive learning.
   \item Through extensive experiments, we demonstrate our network outperforms several alternative methods with clear margins.
\end{itemize}

\section{Related Work}

\subsection{Personal Fashion Recommendation}
Personalization is vital to all online selling services~\citep{mck}. Some work focus on recommending items based on the user preference~\citep{chen2019pog,liu2018personal,packer2018visually,lu2019learning,li2020hierarchical,sagar2020pai}, \eg purchasing records,  social media posts~\citep{zheng2021personalized,sonie2019personalised}, \eg information from Instagram, or body shape~\citep{hidayati2018dress}. Specifically, Packer \etal proposed an approach to personalize clothing recommendation that models the dynamics of individual users' visual preferences by using interpretable image representations generated with a unique feature learning process~\citep{packer2018visually}. Wen \etal. constructed knowledge graphs of user, clothing, and context, and utilized the Apriori algorithm to capture the intrinsic correlations between clothing attributes and context attributes~\citep{wen2018personalized}. Chen \etal. connected user preferences regarding individual items and outfits with Transformer architecture~\citep{chen2019pog}. Zheng \etal. presented an item-to-set metric learning framework that learns to compute the similarity between a set of historical fashion items of a user to a new fashion item~\citep{zheng2021personalized}. Kim \etal. proposed a KD framework for outfit recommendation, which exploits false-negative information from the teacher model while not requiring the ranking of all candidates~\citep{kim2021false}.
Different from the above personal fashion recommendation, this work defines a new task, w.r.t., Fashion Cognitive Learning, which learns the compatibility among outfits and varied personal physical information. ~\citep{hidayati2018dress} learned the compatibility of clothing styles and body shapes which belongs to the scope of this new task. However, they only focused on one aspect \ i.e., body shape of personal physical features. In addition, the way they construct the dataset is to crawl images of stylish female celebrities by issuing each stylish celebrity name combined with a clothing category to be collected as a search query. We follow the common practice of fashion compatibility learning to build the O4U dataset with varied labels of personal features.

\begin{table}[t]
   \begin{center}
   \small{}
   \setlength{\tabcolsep}{1mm}
   \caption{Details of personal physical features and their sub-features.
   }
   \label{physical_label}
   \begin{tabular}{l|c}
   \hline
   \textbf{Features}&\textbf{Sub-features} (N - numbers of sub-features)\\
   \hline
   \hline
   Body Shape&rectangle, top hourglass, athletics, diamond, round, spoon, bottom hourglass... (10)\\
   Skin Color&yellow, dark, fair, brown (4)\\
   Hair Style&long curls, long straight hair... (6)\\
   Hair Color&ginger,black, dark brown, light brown... (6)\\
   Height&high, middle, low (3)\\
   Breasts Size&big, average, small (3)\\
   Color-contrast&high, low (2)\\
   \hline
   \end{tabular}
   \end{center}
   \vspace{-1em}
   \end{table}
   \setlength{\tabcolsep}{1.4pt}

\begin{figure}[t]
   \begin{center}
   \includegraphics[width=1.0\linewidth]{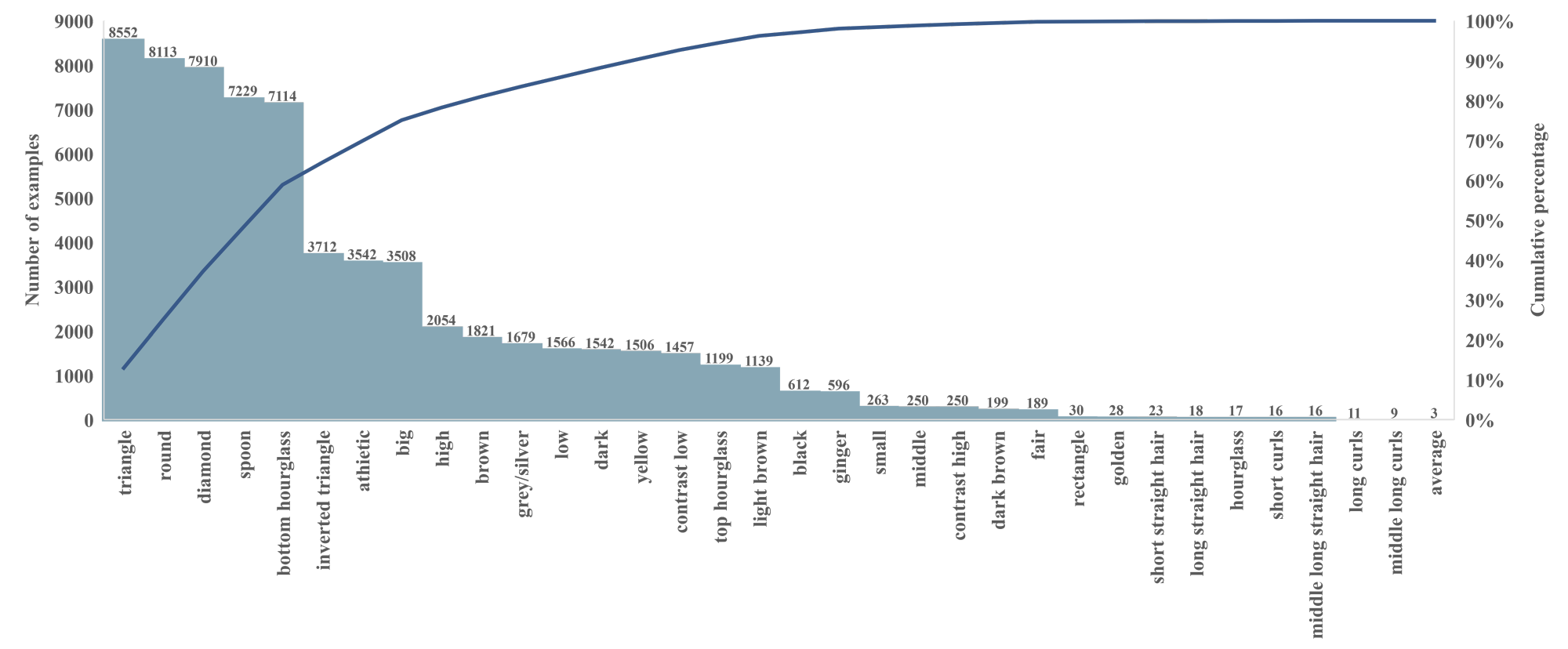}
   \vspace{-1cm}
   \end{center}
   \caption{Number of examples for each physical label}
   \label{dataset}
   \vspace{-0.5cm}
\end{figure}

\section{O4U Dataset}
Fashion cognitive learning is based on fashion compatibility learning to further learn the compatibility among outfits and personal physical features. Thus, we build the Outfit for You (O4U) dataset following the same structure of fashion compatibility learning.

Firstly, we created a total of 29,352 outfits consisting of at least clothing items covering the whole body, one bag, one pair of shoes, and $n$ accessories ($n \in [0,5]$). Then, we invited six experts majoring in fashion to label those outfits. If the outfit is good, they will further select which personal features are not compatible with this outfit. Otherwise, the outfit only has a label to indicate that it is not well-matched. The details of defined personal features are summarized in Table~\ref{physical_label}. There are 15,748 outfits that are considered as good and have an average of 5.25 unmatched physical labels. We randomly divided the dataset into a training set, validation set, and test set in the form of 8:1:1. The label distribution of the training set is shown in Figure~\ref{dataset}. It can be seen that as the OU4 dataset is highly imbalanced, we will introduce our solutions in the following Approach section.

\begin{figure*}[t]
  \begin{center}
  \includegraphics[width=1.0\linewidth]{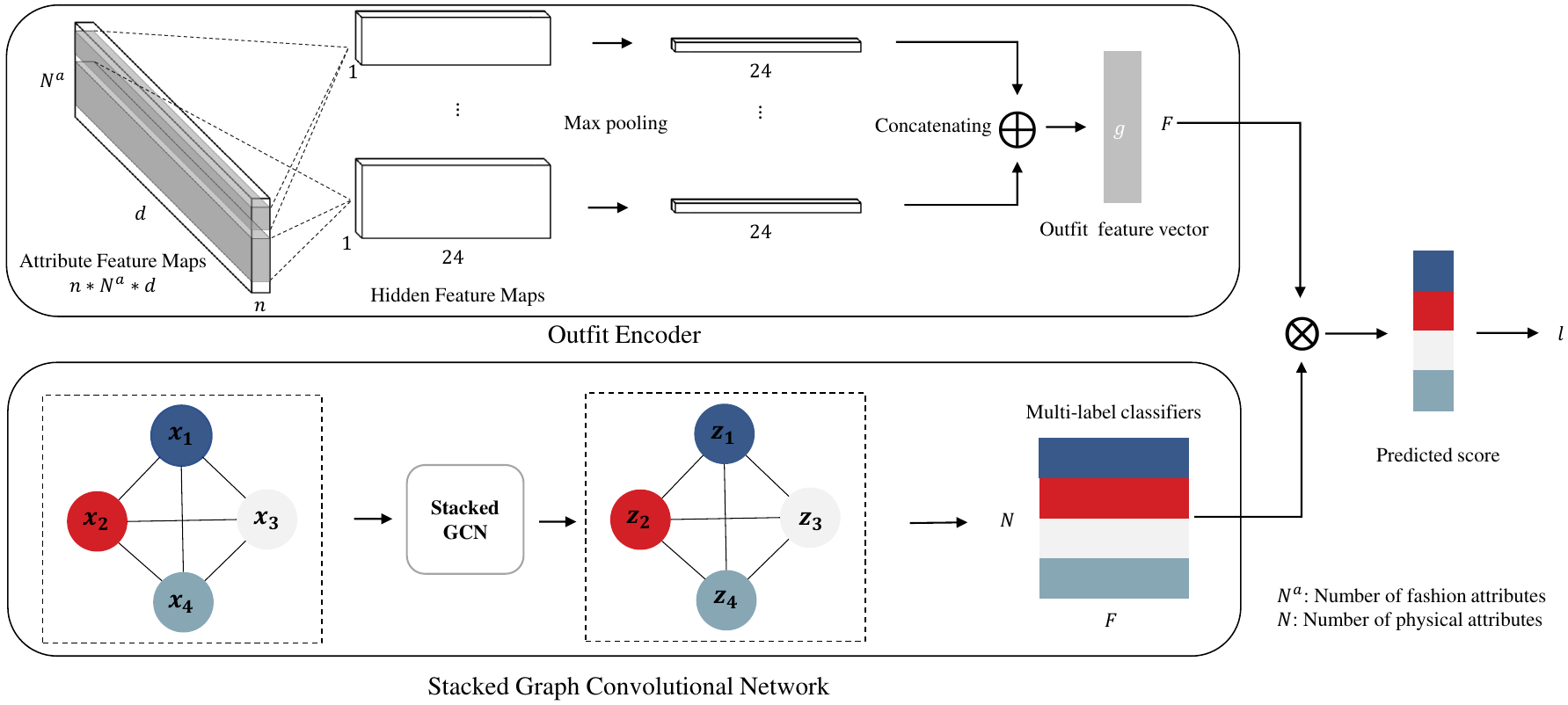}
  \vspace{-1cm}
  \end{center}
  \caption{The framework of Fashion Convolutional Network}
  \label{FCN}
  \vspace{-0.5cm}
\end{figure*}
\section{Approach}
\subsection{Problem Formulation and Motivation}
We define a new task, w.r.t. Fashion Cognitive Learning, which aims to learn the compatibility among outfits and personal physical information. Specifically, given a set of items $\mathcal{M}=\{p_i\}_i^{N_p}$ of $N_p$ individual items and a collection  $\mathcal{T}=\{O_j\}_{j=1}^{N_t}$ containing $N_t$ outfits, where each outfit $O=\{p_i\}_i^n$ in collection $\mathcal{T}$ is defined as a subset of $\mathcal{M}$ containing $n$ different items.
Each outfit $O$ has a fashion compatibility label $l_f \in \{0, 1\}$ indicating whether this outfit is good matched or not and a set of personal physical labels $\mathbf{l_p} \in \mathbb{R}^{N}$, where $N$ is the number of physical labels defined in Table~\ref{physical_label}.
Each item $p_i \in \mathcal{M}$ has its corresponding image $I_i$ (unstructured data) and other metadata such as the primary color data, the category label, and some attribute labels $\mathbf{l_a}$ (structured data). 
Fashion cognitive learning is defined as a multi-label classification task that aims to recognize the given outfit whether it is compatible with some physical labels.

How to encode an outfit into a meaningful embedding is crucial for fashion cognitive learning. In this work, we propose to use 1-dimensional convolutional filters of different sizes to extract the hidden features of the outfit. The motivation for using a convolutional structure to encode outfits is mainly twofold: 1. we observed that there exists translation equivalence in outfit data. In other words, if we swap the order of items or attributes, the outfit embedding should remain the same. Data with such characteristics is suitable to be learned through convolutional structure; 2. we observed that whether an outfit is compatible with a physical label depends on one or several fashion attributes. Using different size convolutional filters can group attribute features in the different numbers together, and that improves the continuity of using the outfit data.

\subsection{Fashion Convolutional Network}
We propose the Fashion Convolutional Network (FCN) to address the personalized recommendation task in terms of human physical labels. FCN includes two modules namely outfit encoder and stacked graph neural networks. 

\subsubsection{Outfit Encoder}
To better reveal the continuity of outfit data, we propose to use a convolutional network to encode an outfit into an outfit embedding, as shown in Figure~\ref{FCN}. Different from using a conventional convolutional neural network (CNN) to extract features from item images, our proposed outfit encoder is applied to attribute features. Given an outfit, fashion attribute features $\mathbf{X} \in \mathbb{R}^{N_a \times d}$ is extracted from each item using well-pretrained VGG~\citep{simonyan2014very} network on a large-scale fashion attribute dataset, where $N_a$ is the number of fashion attributes and $d$ is the dimensionality of attribute features. Each attribute feature vector is the output of the last convolution layer after a max-pooling operation. These attribute features, serving as the input of the outfit encoder, are fixed during the whole training process. The advantage of using these attribute features rather than raw images is that the network can focus on the important features of items and make the training process more efficient.

Attribute feature maps of the given outfit with $n$ items is presented by stacking (padded where necessary) all attribute features along the item dimension,
\begin{equation}
   \mathbf{Z} = \mathbf{X}_1 \oplus \mathbf{X}_2 \oplus \dots \oplus \mathbf{X}_n
\end{equation}
where $\mathbf{Z} \in \mathbb{R}^{n \times N_a \times d}$ and $\oplus$ is the stacking operation.

A convolutional layer contains $N_c$ convolutional filters with different window sizes, and each filter has multiple convolutional kernels. We use the notation $\mathbf{w}_j \in \mathbb{R}^{h_j \times d}$ to denote $j$-th filters in this layer, where $h_j$ means the filter is applied to a window of $h_j$ attribute features to generate a new feature. The input channels of this convolutional layer are $n$ with output channels being 24 for each filter. The convolution stride and padding are fixed to 1 and 0, respectively. After the convolutional process, a max-pooling layer along the filter moving dimension is applied, yielding a 24-dimensional vector for each filter. The final outfit embedding is obtained by concatenating these convolved vectors and is denoted by $g \in \mathbb{R}^{F}$, where $F$ is the dimensionality of the outfit embedding.

\subsubsection{Multi-label Graph Convolutional Network}
We use Multi-label Graph Convolutional Network~\citep{chen2019multi} (ML-GCN) to train classifiers of the physical labels. ML-GCN is a graph convolutional networks (GCN)~\citep{kipf2016semi} based model which takes advantage of capturing the label correlations by treating the classifiers of labels as nodes. We briefly describe how we apply ML-GCN in this paper.
The layer-wise propagation rule of a GCN layer is:
\begin{equation}
    \label{prop}
    x^{(l+1)}=\sigma(\widetilde{D}^{-\frac{1}{2}}\widetilde{A}\widetilde{D}^{-\frac{1}{2}} x^{(l)} \Theta^{(l)} )
\end{equation}
where $\widetilde{A}=A+I_D$ is the adjacency matrix of the graph with self-connections and $\widetilde{D}=\sum_{j}\widetilde{A}_{ij} $ is the degree matrix.
$A \in \mathbb{R}^{N \times N}$ is the adjacency matrix where $N$ denotes the number of nodes in the graph.
$x^{(l)} \in \mathbb{R}^{N \times C^{(l)}}$ is the matrix of activations in the $l^{th}$ layer with $C^{(l)}$ feature maps.
$\Theta^{(l)} \in \mathbb{R}^{C^{(l)} \times C^{(l+1)}}$ is the trainable weight matrix.
$\sigma(\cdot)$ denotes the nonlinear activation function.
$x^{(l+1)} \in \mathbb{R}^{N \times C^{(l+1)}}$ is the convolved feature matrix with $C^{(l+1)}$ feature maps.

We use two-layer stacked GCN to learn classifiers using the layer-wise propagation rule of Eq.~\ref{prop}. Taking the label representation with $C$ physical labels $X \in \mathbb{R}^{N \times C}$ and the adjacency matrix $A \in \mathbb{R}^{N \times N}$ as input, the model $f(X,A)$ can be expressed mathematically as:
\begin{equation}
    Z=f(X,A) = \mathrm{softmax} (\widehat{A} \ \mathrm{ReLU} (\widehat{A}XW^{(0)})W^{(1)})
\end{equation}
where $\widehat{A}=\widetilde{D}^{-\frac{1}{2}}\widetilde{A}\widetilde{D}^{-\frac{1}{2}}$ is normalized version of adjacency matrix. $W^{(0)} \in \mathbb{R}^{C \times H}$ and $W^{(1)} \in \mathbb{R}^{H \times F}$ are two trainable weight matrices for the first and second layer, respectively and $H$ is the dimension of the hidden layer. $Z \in \mathbb{R}^{N \times F}$ is the classifier matrix with $F$ feature maps. For the construction of $A$, every entry in this matrix $A_{ij}=P(L_j | L_i)$ is the conditional probability of label $L_j$ when label $L_i$ appears. So the adjacency matrix is a weighted, asymmetrical matrix.

Given the outfit embedding $\mathbf{g} \in \mathbb{R}^{F}$, the predicted score is the product of label classifiers and the outfit embedding:
\begin{equation}
    \widehat{\mathbf{y}} = Z \cdot g
\end{equation}

\subsubsection{Objective Function}
We evaluate the multi-label classification loss as follows:
\begin{equation}
    L_1=\sum_{n=1}^{N}\mathbf{y}^n \ \textrm{log}(\sigma (\widehat{\mathbf{y}}^n))+(1-\mathbf{y}^n) \ \textrm{log}(1-\sigma (\widehat{\mathbf{y}}^n))
\end{equation}
where $\mathbf{y} \in \mathbb{R}^N$ is ground truth physical labels of an outfit, and $\sigma(\cdot)$ is the sigmoid function.
The overall cost function is defined as follows:
\begin{equation}
   J(\Theta_\mathrm{FCN} ) = L_1 + \frac{\lambda }{2} \left \| \Theta_\mathrm{FCN} \right \| _2^2
\end{equation}
where $\Theta_\mathrm{FCN}$ is the trainable parameters of FCN and $\lambda$ is L2 regularization hyperparameter.

\section{Experiment}
\noindent\textbf{Implementation Details.}
For the outfit encoder, the convolutional layer has five filters with different window sizes, which are 1, 2, 4, 6, and 8. The number of fashion attributes $N^a=14$. For the GCN module, we use a two-layer stacked GCN with each output dimension 200 and 120, respectively. 
The pretrained \textit{VGG}~\citep{simonyan2014very} was used as the attribute feature extractor. Both the height and width of the images were cropped to 224, and the dimensionality of the attribute feature vector is 512. The main color data extracted by FOCO system~\citep{zou2019foco} was also added to the input feature maps. The physical label was encoded by Glove~\citep{pennington2014glove} into 100-dimensional word embeddings.
The FCN was trained in an end-to-end manner on the O4U dataset with a batch size of 10 on NVIDIA RTX 3070 GPU. We use SGD~\citep{Robbins2007ASA} as the optimizer with the learning rate, momentum, and weight decay are $1e^{-1}$, 0.9, and 5e-5, respectively. An exponential decreasing schedule for the learning rate and an early stop training strategy was adopted. 

\noindent\textbf{Compared Approaches.}
\noindent\textbf{1. SVM~\citep{scikit-learn}:}
We use the support vector machine (SVM) as one of our baselines to demonstrate the effectiveness of our approach. We run the SVM classifier provided by scikit-learn on the default setting.
\noindent\textbf{2. Linear:}
A network consisting of multiple linear layers.
\noindent\textbf{3. ResNet~\citep{he2016deep}:}
We trained the ResNet with the input of the mean value of all item images.
\noindent\textbf{4. Attention~\citep{vaswani2017attention}:}
We applied several stacked multi-head attention layers to encode an outfit with various attribute vectors into one vector.

\subsection{Quantitative Results}
\begin{table*}[t]
   \begin{center}
      \small{}
      \setlength{\tabcolsep}{1mm}
      \caption{Quantitative results on Body Shape attributes.}
      \label{ap1}
      \begin{tabular}{l|c|c|c|c|c|c|c|c}
      \hline
      Methods&top hourglass&hourglass&athletics&inverted triangle&triangle&spoon&round&dimension\\
      \hline
      Linear&\textbf{15.47}&63.90&66.47&76.14&63.41&63.09&71.96&70.90\\
      ResNet~\citep{he2016deep}&10.73&31.76&33.37&79.22&67.86&67.15&66.44&65.26\\
      Attn~\citep{vaswani2017attention}&9.48&30.20&31.69&69.68&59.07&57.46&61.36&61.52\\
      \textbf{FCN}&15.39&\textbf{66.53}&\textbf{70.15}&\textbf{83.48}&\textbf{70.29}&\textbf{69.82}&\textbf{77.52}&\textbf{76.35}\\
      \hline
      \hline
   \end{tabular}
   \end{center}
   \vspace{-0.5cm}
\end{table*}

\begin{table*}[t]
   \begin{center}
      \small{}
      \setlength{\tabcolsep}{1mm}
      \caption{Quantitative results on the rest attributes, excepting Body Shape.}
      \label{ap2}
      \begin{tabular}{l|c|c|c|c|c|c|c|c|c}
      &\multicolumn{3}{c|}{Skin}&\multicolumn{2}{c|}{Hair color}&\multicolumn{2}{c|}{Height}&{Breasts}&{Contrast}\\
      \hline

      Methods&yellow&dark&brown&light brown&grey&high&low&big&low\\
      \hline
      Linear&11.59&24.35&14.35&9.17&13.31&17.38&13.94&31.23&12.27\\
      ResNet~\citep{he2016deep}&12.05&41.57&14.29&\textbf{9.83}&\textbf{13.68}&18.08&11.98&26.50&12.06\\
      Attn~\citep{vaswani2017attention}&12.31&11.68&14.27&8.42&12.24&14.30&12.43&27.51&12.29\\
      \textbf{FCN}&\textbf{13.24}&\textbf{46.84}&\textbf{15.11}&9.31&13.00&\textbf{21.57}&\textbf{23.23}&\textbf{31.91}&\textbf{12.72}\\
      \hline
      \hline
   \end{tabular}
   \end{center}
   \vspace{-0.5cm}
\end{table*}

\begin{table*}[t!]
   \begin{center}
      \small{}
      \setlength{\tabcolsep}{0.8mm}
      \caption{Main metrics on 17 physical attributes.}
      \label{main_indi}
      \begin{tabular}{l|c|c|c|c|c|c|c||c|c|c|c|c|c}&{}
      &\multicolumn{6}{c||}{All}&\multicolumn{6}{c}{Top-3}\\
      \hline
      Methods&mAP&CP&CR&CF1&OP&OR&OF1&CP&CR&CF1&OP&OR&OF1\\
      \hline
      SVM~\citep{scikit-learn}&-&28.07&33.10&30.38&68.70&61.54&64.90&-&-&-&-&-&-\\
      Linear&37.59&26.59&\textbf{33.93}&29.81&63.23&\textbf{65.14}&64.17&28.96&20.57&24.06&68.25&41.29&51.46\\
      ResNet~\citep{he2016deep}&34.22&22.83&27.55&24.97&64.29&57.18&60.53&23.98&18.80&21.08&67.52&40.06&50.29\\
      Attn~\citep{vaswani2017attention}&29.76&18.18&29.41&22.47&61.82&62.33&62.07&11.44&17.65&13.88&64.82&39.22&48.87\\
      \hline
      \textbf{FCN}&\textbf{42.14}&\textbf{32.29}&33.84&\textbf{33.04}&\textbf{68.89}&62.17&\textbf{65.36}&\textbf{34.16}&\textbf{21.06}&\textbf{26.06}&\textbf{73.25}&\textbf{41.32}&\textbf{52.83}\\

      \hline
      \hline
   \end{tabular}
   \end{center}
   \vspace{-0.5cm}
\end{table*}

\noindent\textbf{Evaluation metrics.}
Following the general practice~\citep{wang2016cnn, chen2019multi, tarekegn2021review}, we report the performance of models on these metrics: mean average precision (mAP); average per-class precision (CP), recall (CR), and F1 (CF1); average overall precision (OP), recall (OR), and F1 (OF1). Average per-class metrics will first evaluate each label individually and then average over all labels. Average overall metrics will evaluate over all examples. We also report the results of these metrics on top-3 labels.

We report mAP results for 17 physical labels in Tables~\ref{ap1} and~\ref{ap2}. Our proposed method FCN achieves the best performance over 14 out of 17 labels compared with other baseline methods. Especially on labels belonging to the body shape category, our method achieves a huge improvement compared to other methods.

Model performance covering all labels is reported in Table~\ref{main_indi}. FCN outperforms other baseline methods on all metrics. SVM, an effective machine learning method, shows good performance in terms of average overall metrics. However, FCN improves by 4.22, 0.74, and 2.66 on the CP, CR, and OP. The performance of ResNet is not good on mAP, and it indicates that treating outfits as the mean value of item images is not a good idea for this task. As for the method using the attention mechanism, although it outperforms FCN on the OR metric, the value of mAP is much lower than FCN.

\subsection{Qualitative Results}
\begin{figure*}[t]
   \begin{center}
   \includegraphics[width=1.0\linewidth]{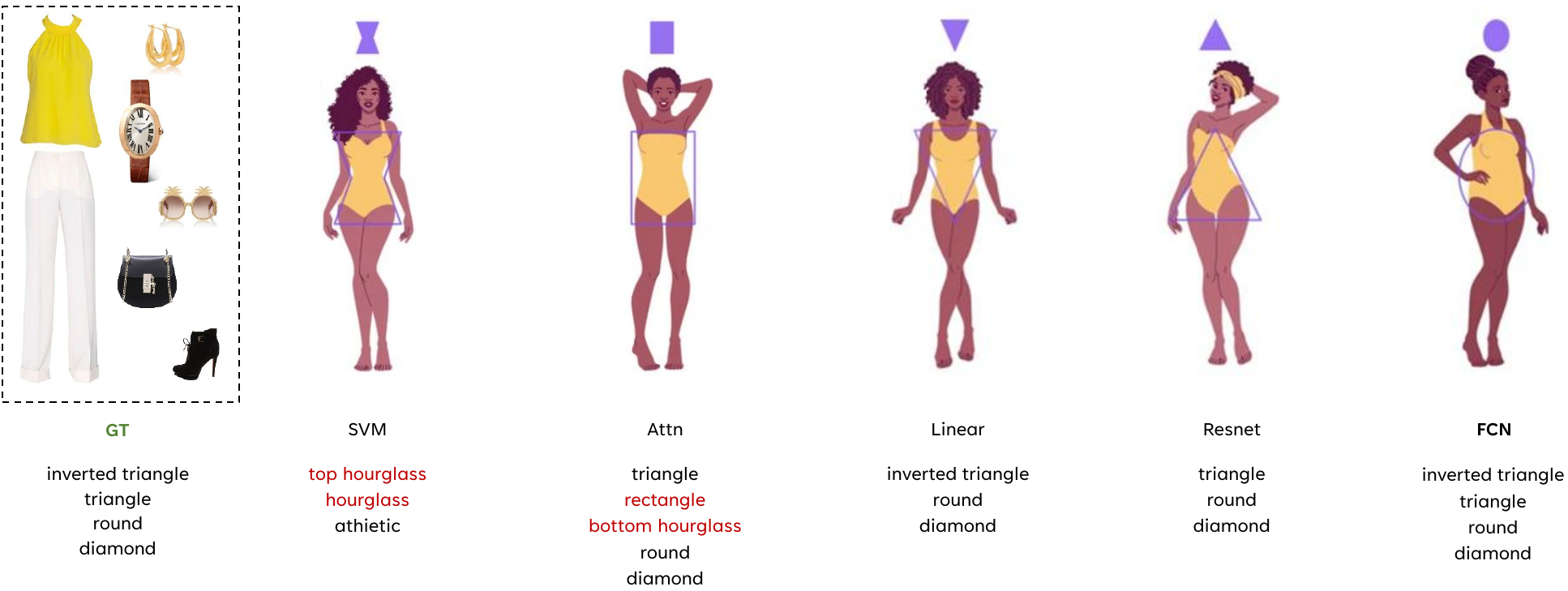}
   \vspace{-1cm}
   \end{center}
   \caption{Qualitative results of all compared methods and the proposed FCN.}
   \label{q}
   \vspace{-0.5cm}
\end{figure*}
Additionally, we present the qualitative results in Figure~\ref{q}. People with body figures including inverted triangle, triangle, round, diamond is not suitable for the outfit in the left side. The main reason is that the silhouette of the tank top and the straight-line pants are not matched with these types of body shape. It can be seen that the FCN accurately points out all incompatible body shape among the comparative methods and thus indicates that our method well learns the compatibility among fashion outfit and personal physical information. 

\begin{table*}[t!]
   \begin{center}
      \small{}
      \setlength{\tabcolsep}{0.6mm}
      \caption{Effect of filter region size.}
      \label{region_size}
      \begin{tabular}{l|c|c|c|c|c|c|c||c|c|c|c|c|c}&{}
         &\multicolumn{6}{c||}{All}&\multicolumn{6}{c}{Top-3}\\
         \hline
         Region size&mAP&CP&CR&CF1&OP&OR&OF1&CP&CR&CF1&OP&OR&OF1\\
         \hline
         (1)&40.68&32.55&32.99&32.77&67.68&62.50&64.99&29.56&20.27&24.05&71.29&40.53&51.67\\
         (2)&38.93&28.36&32.42&30.25&68.67&61.28&64.77&30.64&20.44&24.52&73.01&40.24&51.89\\
         (4,4,4,4,4)&\textbf{43.11}&\textbf{32.82}&33.46&\textbf{33.13}&68.70&62.02&65.19&32.30&20.82&25.32&72.30&40.87&52.22\\
         (8,9,10)&41.38&28.29&32.72&30.35&68.83&61.29&64.85&30.29&21.19&24.93&73.12&40.96&52.51\\
         (1,2,4,6,8)&42.14&32.29&\textbf{33.84}&33.04&\textbf{68.89}&\textbf{62.17}&\textbf{65.36}&\textbf{34.16}&\textbf{21.06}&\textbf{26.06}&\textbf{73.25}&\textbf{41.32}&\textbf{52.83}\\
         \hline
         \hline
      \end{tabular}
   \end{center}
   \vspace{-0.5cm}
\end{table*}

\begin{table*}[t!]
   \begin{center}
      \small{}
      \setlength{\tabcolsep}{0.6mm}
      \caption{Effect of numbers of kernels for each filter.}
      \label{kernels}
      \begin{tabular}{l|c|c|c|c|c|c|c||c|c|c|c|c|c}&{}
         &\multicolumn{6}{c||}{All}&\multicolumn{6}{c}{Top-3}\\
         \hline
         No. Kernels&mAP&CP&CR&CF1&OP&OR&OF1&CP&CR&CF1&OP&OR&OF1\\
         \hline
         2&35.46&26.54&35.58&30.40&62.52&68.25&65.26&27.94&19.72&23.12&66.03&39.95&49.78\\
         12&41.67&32.19&33.91&33.03&67.96&63.09&65.44&\textbf{36.09}&20.77&\textbf{26.37}&72.31&40.95&52.29\\
         24&42.14&32.29&\textbf{33.84}&\textbf{33.04}&68.89&\textbf{62.17}&\textbf{65.36}&34.16&\textbf{21.06}&26.06&73.25&\textbf{41.32}&\textbf{52.83}\\
         48&\textbf{42.65}&\textbf{32.55}&33.10&32.82&\textbf{69.17}&60.90&64.77&34.75&21.02&26.20&{73.75}&40.78&52.52\\
         \hline
         \hline
      \end{tabular}
   \end{center}
   \vspace{-0.5cm}
\end{table*}

\begin{table*}[t!]
   \begin{center}
      \small{}
      \setlength{\tabcolsep}{0.7mm}
      \caption{Effect of numbers of GCN layers.}
      \label{effect_gcn}
      \begin{tabular}{l|c|c|c|c|c|c|c||c|c|c|c|c|c}&{}
         &\multicolumn{6}{c||}{All}&\multicolumn{6}{c}{Top-3}\\
         \hline
         No. GCN&mAP&CP&CR&CF1&OP&OR&OF1&CP&CR&CF1&OP&OR&OF1\\
         \hline
         1&40.85&\textbf{34.11}&32.09&\textbf{33.07}&68.19&61.38&64.61&25.01&18.93&21.55&71.63&39.67&51.06\\
         2&\textbf{42.14}&32.29&\textbf{33.84}&33.04&68.89&\textbf{62.17}&\textbf{65.36}&\textbf{34.16}&\textbf{21.06}&\textbf{26.06}&\textbf{73.25}&\textbf{41.32}&\textbf{52.83}\\
         4&40.73&28.59&32.24&30.31&\textbf{69.05}&60.46&64.47&30.49&20.83&24.75&73.02&40.40&52.02\\
         8&39.45&28.00&32.29&29.99&67.73&60.19&63.74&21.25&19.55&20.36&71.18&35.54&47.41\\
         \hline
         \hline
      \end{tabular}
   \end{center}
\end{table*}

\subsection{Ablation Study}
\noindent\textbf{Effect of filter region size.}
We explore the sensitivity of different combinations of filter region size. As shown in Table~\ref{region_size}, only using one kind of convolutional filter size shows the worst performance. Using filters with a big region size (relative to attribute number 14) has a negative effect on model performance. Using multiple filters with the same size achieves the best result on mAP and OF1, but the results are lower than FCN on the top-3 labels. The combination we use in FCN (1, 2, 4, 6, 8) shows the best performance on CF1 and Top-3 metrics.

\noindent\textbf{Effect of numbers of kernels for each filter}
We then explore the effect of the different numbers of kernels while keeping the filter region size to be the same and report the results in Table~\ref{kernels}. We find that the performance achieves the best results when the number of kernels is 24. Using too few convolutional kernels will deteriorate performance significantly. Using too many kernels cannot dramatically improve performance while it causes a negative impact on recall metrics.

\noindent\textbf{Effect of numbers of GCN layers}
We report the effects of different numbers of GCN layers in Table~\ref{effect_gcn}. We found that with deeper multi-layer GCNs, the performance degrades on almost all metrics. Therefore we choose to use a two-layer stacked GCN in our model.

\section{Conclusion}
We introduce a new task, Fashion Cognitive Learning, which targets to learn the compatibility among fashion outfits and personal physical information. The FCN, which contains two submodules that are outfit encoder and Multi-label Graph Neural Network (ML-GCN), is proposed to learn the relationships among visual-semantic embedding of outfit composition and appearance features of individuals. 
For implementation, we build a new large-scale fashion outfit dataset, O4U, covering comprehensive personal physical information. Extensive results demonstrate the advance of the proposed framework compared with all alternative methods.

\bibliography{egbib}
\end{document}